\documentclass[11pt]{article}
\usepackage[hyperref]{acl}
\usepackage{times}
\usepackage{latexsym}
\usepackage{amsmath}
\usepackage{amssymb}
\usepackage{graphicx}
\usepackage{booktabs}
\usepackage{multirow}
\usepackage{xcolor}
\usepackage{pifont}
\usepackage{url}

\newcommand{\Ls}{\lambda_{\mathrm{r}}}
\newcommand{\Hv}{H(\mathbf{v})}
\newcommand{\hvec}{\mathbf{h}}
\newcommand{\wvec}{\mathbf{w}}

\title{Whitening Reveals Cluster Commitment as the Geometric Separator of Hallucination Types}

\author{
  Matic Korun \\
  Independent Researcher \\
  Ljubljana, Slovenia \\
  \texttt{iam.m3x@gmail.com}
}

\begin{document}
\maketitle


\begin{abstract}
A geometric hallucination taxonomy distinguishes three failure types---center-drift (Type~1), wrong-well convergence (Type~2), and coverage gaps (Type~3)---by their signatures in embedding cluster space. Prior work found Types~1 and~2 indistinguishable in full-dimensional contextual measurement. We address this through PCA-whitening and eigenspectrum decomposition on GPT-2-small, using multi-run stability analysis (20 seeds) with prompt-level aggregation. Whitening transforms the micro-signal regime into a space where peak cluster alignment (max\_sim) separates Type~2 from Type~3 at Holm-corrected significance, with condition means following the taxonomy's predicted ordering: Type~2 (highest commitment) $>$ Type~1 (intermediate) $>$ Type~3 (lowest). A first directionally stable but underpowered hint of Type~1/2 separation emerges via the same metric, generating a capacity prediction for larger models. Prompt diversification from 15 to 30 prompts per group eliminates a false positive in whitened entropy that appeared robust at the smaller set, demonstrating prompt-set sensitivity in the micro-signal regime. Eigenspectrum decomposition localizes this artifact to the dominant principal components and confirms that Type~1/2 separation does not emerge in any spectral band, rejecting the spectral mixing hypothesis. The contribution is threefold: whitening as preprocessing that reveals cluster commitment as the theoretically correct separating metric, evidence that the Type~1/2 boundary is a capacity limitation rather than a measurement artifact, and a methodological finding about prompt-set fragility in near-saturated representation spaces.
\end{abstract}


\section{Introduction}

Can a 124-million-parameter language model distinguish between drifting under weak context and committing to the wrong semantic region? The question matters because hallucination detection increasingly relies on internal representations \citep{tonmoy2024comprehensive, zhang2023siren, kadavath2022language}, and recent work suggests that models encode hallucination-relevant information in their hidden states \citep{duan2024llmsknow}, yet the geometric structure of these representations under different failure modes remains poorly understood. Companion work \citep{author2026induction} found that coverage-gap hallucinations (Type~3) produce geometrically distinctive contextual representations at prompt level, but center-drift (Type~1) and wrong-well convergence (Type~2) remained indistinguishable in full-dimensional measurement. Two hypotheses were proposed: a capacity limitation at 124M parameters, or spectral mixing that dilutes a band-specific signal when full-dimensional metrics aggregate across all principal components.

This paper resolves the question through PCA-whitening with multi-run stability analysis (20 independent generation seeds, $N = 30$ prompts per group), supplemented by eigenspectrum band decomposition. The results reframe the problem: the theoretically correct separating metric is not entropy $\Hv$ but peak cluster alignment (max\_sim)---the metric that directly measures the taxonomy's defining property of cluster commitment.

The key findings are: (1)~whitened max\_sim separates Type~2 from Type~3 at 40\% Holm-corrected significance ($r = -0.31$, direction 20/20), with condition means following the predicted ordering $\text{T2} > \text{T1} > \text{T3}$; (2)~the first hint of Type~1/2 separation appears via the same metric (15\% Holm, $r = +0.21$, direction 17/20), underpowered at 124M but directionally stable---generating a capacity prediction; (3)~whitened entropy, which appeared to be the strongest result at $N = 15$, is a prompt-specific artifact eliminated by diversification to $N = 30$; (4)~eigenspectrum decomposition localizes this artifact to PCs~1--16 and confirms no Type~1/2 separation in any spectral band, rejecting the spectral mixing hypothesis.


\section{Background}

\subsection{The Three-Type Taxonomy}

The hallucination taxonomy classifies generation failures by their geometric relationship to the ambient cluster structure of token embeddings \citep{ji2023survey, huang2023survey, su2023characterizing}. \textbf{Type~1 (center-drift):} under weak context, generation drifts toward the embedding centroid, producing low cluster membership entropy $\Hv$ \citep{shannon1948mathematical} and low norm. \textbf{Type~2 (wrong-well):} the model commits to a locally coherent but contextually wrong cluster, producing high $\Hv$ with trajectory discontinuities. \textbf{Type~3 (coverage gap):} the query requires absent semantic combinations, producing weak membership across all clusters.

Paper~1 \citep{author2026geometry} validated the geometric prerequisites ($\alpha$, $\beta$, $\Ls$) across 11 transformer models in static embedding space. Paper~2 \citep{author2026induction} tested whether induced hallucinations produce the predicted signatures in GPT-2-small, complementing benchmark-based evaluation approaches \citep{li2023halueval}, finding: (a)~static embeddings do not encode hallucination type ($p = 0.47$); (b)~contextual hidden-state norm separates conditions at both token and prompt level; (c)~Type~3 separates from Types~1/2 robustly on norm; (d)~Types~1 and~2 do not separate in full-dimensional measurement.

\subsection{The Micro-Signal Regime}

Contextual hidden states in GPT-2 operate in a near-saturated similarity regime: $\Hv \approx 0.985$, max centroid similarity $\approx 0.993$. This anisotropy---where contextual representations cluster in a narrow cone of the ambient space---has been documented across transformer architectures \citep{ethayarajh2019contextual, cai2021isotropy, bis2021too, rajaee2021isotropy}. Meaningful differences between hallucination conditions live in the fourth decimal place of cosine similarity. The detection architecture proposed in the taxonomy paper \citep{author2026geometry}---percentile-based zone thresholds---was designed for the broad dynamic range of static embeddings ($\Hv \approx 0.2$--$0.8$) and fails when applied to this narrow-band regime. Paper~2 identified the need for representational preprocessing to amplify micro-signals into detectable effects.

\subsection{Cluster Commitment as the Separating Property}

The taxonomy's type definitions make a specific prediction about peak cluster alignment: Type~2 (wrong-well) \textit{commits} to a cluster, producing the highest max\_sim; Type~1 (center-drift) drifts without commitment, producing intermediate max\_sim; Type~3 (coverage gap) aligns with no cluster, producing the lowest max\_sim. This ordering---$\text{T2} > \text{T1} > \text{T3}$---is the taxonomy's core geometric prediction, and max\_sim is the metric that directly measures it. Full-dimensional $\Hv$ conflates cluster commitment with distributional spread, potentially masking the distinction. Whitening, by equalizing variance across dimensions, should amplify this metric's discriminative power.

\subsection{Two Hypotheses for the Type~1/2 Collapse}

The non-separation of Types~1 and~2 in Paper~2 admitted two hypotheses:

\paragraph{Capacity hypothesis.} The 124M-parameter GPT-2 lacks the representational precision to encode the difference between weak context (Type~1) and misrouted context (Type~2). This reflects a fundamental asymmetry: coverage gaps (Type~3) involve token combinations outside the training distribution---a distributional anomaly detectable at any scale---while Types~1 and~2 both involve familiar tokens in familiar contexts, differing only in routing quality.

\paragraph{Spectral mixing hypothesis.} The distinction is encoded in specific eigenspectrum bands that are diluted when full-dimensional metrics aggregate across all components. Spectrally targeted measurement should resolve the distinction without scaling.

These hypotheses make opposing predictions. Spectral mixing predicts band-specific separation; the capacity hypothesis predicts absence across all bands.


\section{Experimental Design}

\subsection{Model and Generation}

All experiments use GPT-2-small \citep{radford2019language, vaswani2017attention} (124M parameters, 12 layers, 768D hidden states). Generation uses manual autoregressive decoding with KV cache: at each step, the model processes only the new token while retaining past key--value pairs, and the last-layer last-position hidden state is extracted via \texttt{output\_hidden\_states=True}. This produces one 768-dimensional vector per generated token. All generation uses temperature~1.0 with no top-$k$ or top-$p$ filtering, generating up to 60 tokens per prompt (generation terminates early if an end-of-sequence token is produced).

\subsection{Prompt Design}

The whitened experiment uses $N = 30$ prompts per condition (90 total): 15 per type from Paper~2, plus 15 new prompts per type designed for greater diversity. Type~1 prompts are generic low-constraint starters (``The'', ``For example'', ``According to''). Type~2 prompts exploit lexical polysemy and garden-path constructions (``The bank announced record levels of'', ``The seal was broken on the''). Type~3 prompts target knowledge boundaries through pseudo-academic terminology, contradictions, and absurd compositions (``The xenoplasmic refractometry of late-Holocene'', ``According to the well-established proof that pi is rational''). The expansion from 15 to 30 prompts per type is a critical design choice: it tests whether results observed at $N = 15$ survive prompt diversification.

\subsection{Calibration}

Calibration uses 40 diverse prompts (expanded from 25 in Paper~2) generating up to 60 tokens each (${\sim}2{,}400$ contextual vectors), spanning politics, science, sports, arts, economics, technology, and environment. The calibration distribution defines the whitening transform and cluster structure. Calibration is performed once with a fixed seed (42); only experimental generation varies across runs.

\subsection{Multi-Run Stability Analysis}

Each experiment is repeated with 20 independent generation seeds (1--20). Calibration (background generation, whitening transform, clustering, zone thresholds) is fixed across all runs; only experimental text generation varies. This isolates generation stochasticity as the sole source of run-to-run variation. Statistical tests use prompt-level aggregation ($N = 30$/group for the whitened experiment) as the primary inference level, with token-level results reported as reference only.

For each of 20 runs, we report: prompt-level Mann-Whitney $U$ with rank-biserial effect size $r$ \citep{vargha2000critique}, permutation $p$-values (50,000 permutations or exact enumeration), BCa bootstrap 95\% CIs on $r$ (10,000 resamples), and Holm-Bonferroni correction \citep{holm1979simple} within each metric family. We then aggregate across runs: significance rate (fraction of 20 runs reaching $p < 0.05$), Holm survival rate, median $r$, and directional stability (fraction of runs with consistent sign).

\subsection{Experiment~1: PCA-Whitening}

We compute the calibration mean $\boldsymbol{\mu} \in \mathbb{R}^{768}$ and PCA decomposition. Centering and variance normalization have been shown to improve the discriminative quality of word representations by removing dominant directions that encode frequency rather than meaning \citep{mu2018allbut}. The whitening transform projects centered vectors onto the top 256 principal components (capturing 99.7\% of variance) and scales each by $1/\sqrt{\lambda_i + \epsilon}$ (regularization $\epsilon = 10^{-5}$):
\begin{equation}
\wvec = (\hvec - \boldsymbol{\mu}) \cdot \mathbf{W}, \quad W_{:,i} = \frac{\mathbf{v}_i}{\sqrt{\lambda_i + \epsilon}}
\end{equation}
where $\mathbf{v}_i$ and $\lambda_i$ are the $i$-th eigenvector and eigenvalue. The whitened calibration distribution has identity covariance. We cluster in this space using MiniBatchKMeans \citep{sculley2010web} ($k = 40$, batch size 1024, $n_{\text{init}} = 5$, random state 42) and apply the same zone classification framework as previous experiments.

For each whitened vector, we compute three metrics relative to the 40 cluster centroids: cluster membership entropy $\Hv$ (Shannon entropy of softmax similarities), peak cluster alignment max\_sim (maximum cosine similarity to any centroid), and whitened norm $\|\wvec\|$. Raw (unwhitened) norm is retained as a replication control for Paper~2's contextual results.

\subsection{Experiment~2: Spectral Band Decomposition}

The spectral analysis uses $N = 15$ prompts per group (the original Paper~2 set) due to computational cost (${\sim}24$ hours per 20-seed run on CPU). We compute the full PCA on calibration data (768 components) and define six spectral bands with logarithmic spacing matching eigenvalue decay (Table~\ref{tab:bands}).

\begin{table}[h]
\centering
\scriptsize
\setlength{\tabcolsep}{20pt}
\begin{tabular}{llr}
\toprule
\textbf{Band} & \textbf{PCs} & \textbf{Variance} \\
\midrule
Dominant    & 1--16    & 98.0\% \\
Transition  & 17--48   & 0.7\% \\
Mid-range A & 49--128  & 0.6\% \\
Mid-range B & 129--256 & 0.4\% \\
Lower       & 257--512 & 0.3\% \\
Tail        & 513--768 & $<$0.1\% \\
\bottomrule
\end{tabular}
\caption{Spectral bands defined across the eigenspectrum.}
\label{tab:bands}
\end{table}

For each band, we: (i)~project all vectors onto the band's principal components, (ii)~whiten within that subspace, (iii)~cluster ($k$ adapted to band dimensionality: $\min(40, \text{band\_dim}/2)$, minimum~10), (iv)~compute $\Hv$, max\_sim, and norm, (v)~run two-level statistical tests. Each band is analyzed independently. Given the $\Hv$ false positive identified at $N = 30$ in the whitened experiment, spectral $\Hv$ results in the dominant band should be interpreted as reflecting the prompt-specific artifact rather than a robust signal. The analysis is retained for its diagnostic value in localizing where signal concentrates across the eigenspectrum.


\section{Experiment~1: Whitening}

\subsection{The max\_sim Result}

Whitening transforms the micro-signal regime ($\Hv \approx 0.985$, max\_sim $\approx 0.993$) into a calibrated space where deviations from the background distribution become first-order effects. In this space, max\_sim emerges as the primary separating metric, with condition means following the taxonomy's predicted ordering (Table~\ref{tab:maxsim_results}).

\begin{table}[t]
\centering
\scriptsize
\setlength{\tabcolsep}{7pt}
\begin{tabular}{lrrrrr}
\toprule
\textbf{Pair} & \textbf{Sig rate} & \textbf{Holm} & \textbf{Med $r$} & \textbf{Dir} & \textbf{Pseudo} \\
\midrule
T2--T3 & 60\% & 40\% & $-0.31$ & 20/20 & $0.3\times$ \\
T1--T2 & 35\% & 15\% & $+0.21$ & 17/20 & $1.4\times$ \\
T1--T3 &  5\% &  5\% & $-0.15$ & 18/20$^-$ & $12\times$ \\
\bottomrule
\end{tabular}
\caption{Whitened max\_sim pairwise results ($N = 30$/group, 20 seeds). Sig rate: fraction of runs with prompt-level MW $p < 0.05$. Holm: fraction surviving Holm-Bonferroni correction. Med~$r$: median rank-biserial effect size. Dir: directional stability (runs with consistent sign). Pseudo: pseudoreplication ratio (token sig rate / prompt sig rate); values $< 1$ indicate the prompt-level effect is \textit{stronger} than token-level---a genuine between-condition effect; values $> 1$ indicate token-level inflation (standard pseudoreplication).}
\label{tab:maxsim_results}
\end{table}

The T2--T3 result has the cleanest statistical profile in the trilogy. The pseudoreplication ratio is inverted ($0.3\times$): prompt-level significance (60\%) \textit{exceeds} token-level significance (20\%). This means within-prompt autocorrelation dilutes rather than inflates the effect---the opposite of every other comparison in the paper series. The signal is a genuine between-condition difference that token-level noise actually masks.

Condition means across 20 seeds confirm the predicted ordering: Type~2 ($0.180 \pm 0.004$) $>$ Type~1 ($0.172 \pm 0.004$) $>$ Type~3 ($0.168 \pm 0.003$). Type~2 (wrong-well) shows the highest peak cluster alignment, consistent with committing to a specific semantic region. Type~3 (coverage gap) shows the lowest, consistent with failing to align with any learned cluster. Type~1 (center-drift) falls between, consistent with drifting without commitment.

\subsection{The $\Hv$ False Positive}

At $N = 15$ (the original Paper~2 prompt set), whitened $\Hv$ appeared to be the strongest result across the trilogy. Prompt-level Kruskal-Wallis reached 90\% significance across 20 seeds, with two strong pairwise effects: T1--T3 reached 65\% Holm-corrected significance ($r \approx +0.55$, direction 20/20) and T2--T3 reached 30\% Holm ($r \approx +0.48$, direction 20/20). This was the basis for the original framing of Paper~3. At $N = 30$, both signals collapse: prompt-level Kruskal-Wallis significance drops to 5\%, T1--T3 Holm survival falls to 1/20, and T2--T3 falls to 0/20. The entropy signal was prompt-specific---the original 15 prompts per type happened to produce token distributions that separated on the principal axis of variance (see \S\ref{sec:spectral_localization}). Adding 15 diverse prompts per type broke this alignment.

This collapse is a methodological finding about the micro-signal regime: when meaningful differences live in the fourth decimal place, even moderate prompt-set restriction can produce artifactual separation that appears robust across generation seeds but does not generalize to broader prompt samples.

\subsection{Whitened Norm and Raw Norm Controls}

Whitened norm is effectively destroyed by whitening (5--10\% significance across pairs), as expected: whitening equalizes variance along every principal axis, eliminating magnitude information.

Raw (unwhitened) norm, retained as a replication control, reproduces Paper~2's contextual results with the ordering T1 ($244.7 \pm 2.7$) $>$ T2 ($243.2 \pm 2.3$) $>$ T3 ($238.0 \pm 2.7$). T1--T3 reaches 4/20 Holm (median $r = -0.21$, direction 20/20); T2--T3 reaches 4/20 Holm (median $r = -0.16$, direction 19/20). These are consistent with Paper~2's moderate norm effects at expanded prompt diversity.

\subsection{Type~1/2: The Emerging Signal}

The T1--T2 max\_sim result (35\% sig, 15\% Holm, $r = +0.21$, direction 17/20) is the first empirical hint of Type~1/2 separation in the trilogy. The sign is correct: Type~2 shows higher cluster commitment than Type~1, as the taxonomy predicts. The directional stability (17/20) indicates the sign is real; the low Holm rate indicates the magnitude is insufficient for reliable detection at 124M parameters with 30 prompts. This generates a scaling prediction: at larger model sizes where contextual attractors are sharper, the T1--T2 gap should widen before the T1--T3 gap (which combines the routing distinction with the distributional anomaly of coverage gaps).

\begin{figure*}[t]
\centering
\includegraphics[width=\textwidth]{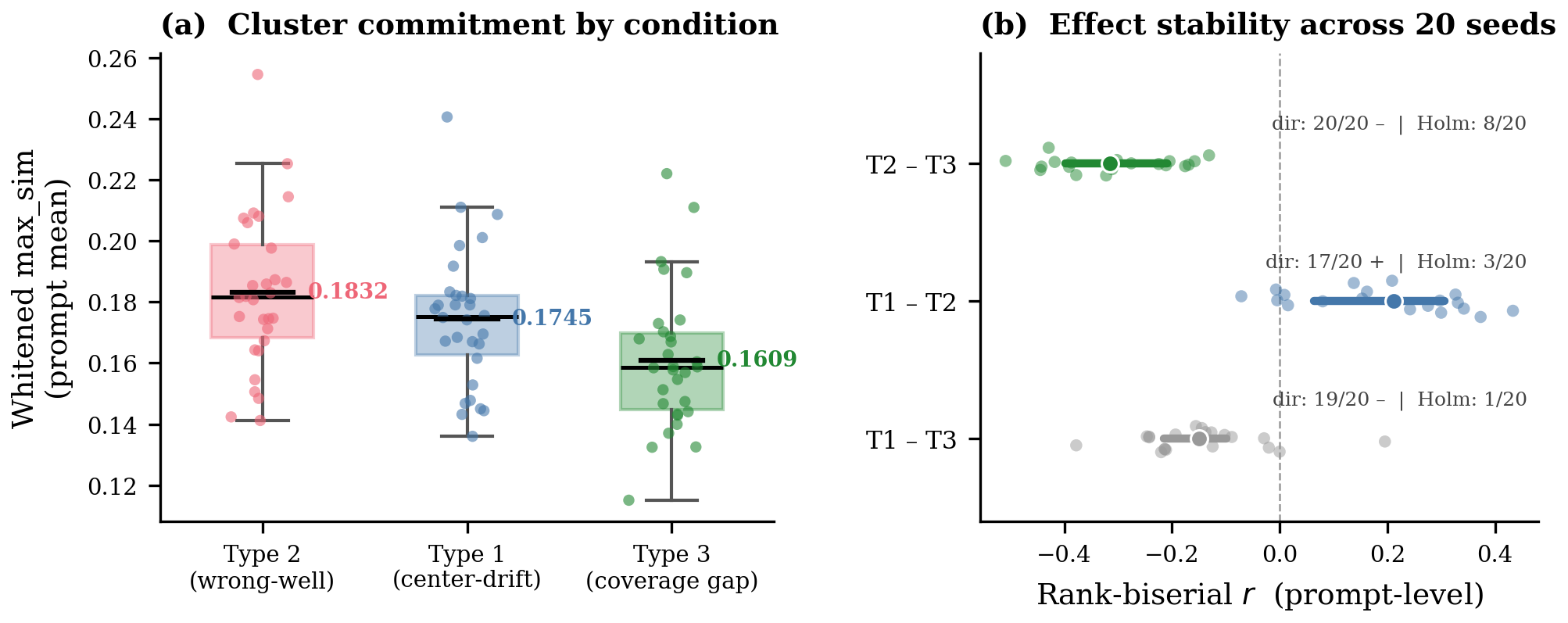}
\caption{Whitened max\_sim results. \textbf{(a)}~Per-prompt mean max\_sim by condition from the representative seed ($N = 30$/group). The ordering T2 $>$ T1 $>$ T3 matches the taxonomy's core prediction: wrong-well commits to a cluster (highest), center-drift drifts without commitment (intermediate), coverage gap aligns with nothing (lowest). Grand means annotated. \textbf{(b)}~Rank-biserial effect size $r$ across 20 independent generation seeds, with median and IQR bars. T2--T3 (green): 20/20 directional stability, 8/20 Holm. T1--T2 (blue): 17/20 directional, 3/20 Holm---the emerging signal. T1--T3 (gray): directionally stable but weak, reflecting the combined gap.}
\label{fig:maxsim_ordering}
\end{figure*}


\section{Experiment~2: Spectral Localization}
\label{sec:spectral_localization}

The spectral analysis ($N = 15$/group, 20 seeds) serves three diagnostic purposes: localizing the $\Hv$ artifact, confirming the absence of hidden mid-range structure, and documenting signal in the tail band.

\subsection{Localizing the $\Hv$ Artifact}

The dominant principal components (PCs~1--16, 98\% of variance) carry the $\Hv$ signal that collapsed at $N = 30$. In this band, $\Hv$ reaches 95\% prompt-level significance for T1--T3 (18/20 Holm, median $r = +0.61$) and 100\% for T2--T3 (20/20 Holm, median $r = +0.75$). These are the strongest prompt-level effects in the entire analysis---and they are the artifact. The spectral view confirms the mechanism: the original 15 prompts per type produced representations that separated along the highest-variance dimensions. Diversification to 30 prompts distributed energy more evenly across the dominant subspace, eliminating the separation.

\subsection{No Hidden Mid-Range Structure}

If the Type~1/2 distinction were spectrally localized, mid-range bands (PCs~49--256) would show separation that the full-dimensional aggregate washes out. They do not. Across PCs~49--128 and PCs~129--256, Type~1/2 prompt-level significance is 5--15\% for all metrics, with Holm survival at 0--10\%. The mid-range is essentially dead for Type~1/2 across all metrics, and largely uninformative for other pairs. This confirms that the spectral mixing hypothesis is rejected not by failure to look in the right place, but by comprehensive absence across the entire eigenspectrum.

\subsection{The Tail Phenomenon}

The tail band (PCs~513--768, $<$0.1\% of variance) shows the strongest prompt-level effects for Type~3 comparisons: max\_sim T2--T3 reaches 90\% significance (16/20 Holm, $r = -0.58$), and norm T2--T3 reaches 90\% significance (16/20 Holm, $r = +0.59$). These are genuine between-condition effects---the pseudoreplication ratios are near 1.0. However, these dimensions contribute negligibly to the actual embedding geometry. The tail signal suggests that GPT-2 encodes type-relevant structure even in its lowest-variance dimensions, but at 124M parameters this structure is not actionable for detection. It becomes a capacity prediction: at larger scale, these dimensions may gain sufficient variance to contribute.

Type~1/2 remains non-significant in the tail (15\% sig, $r \approx +0.23$), consistent with the capacity hypothesis.

\begin{figure*}[t]
\centering
\includegraphics[width=\textwidth]{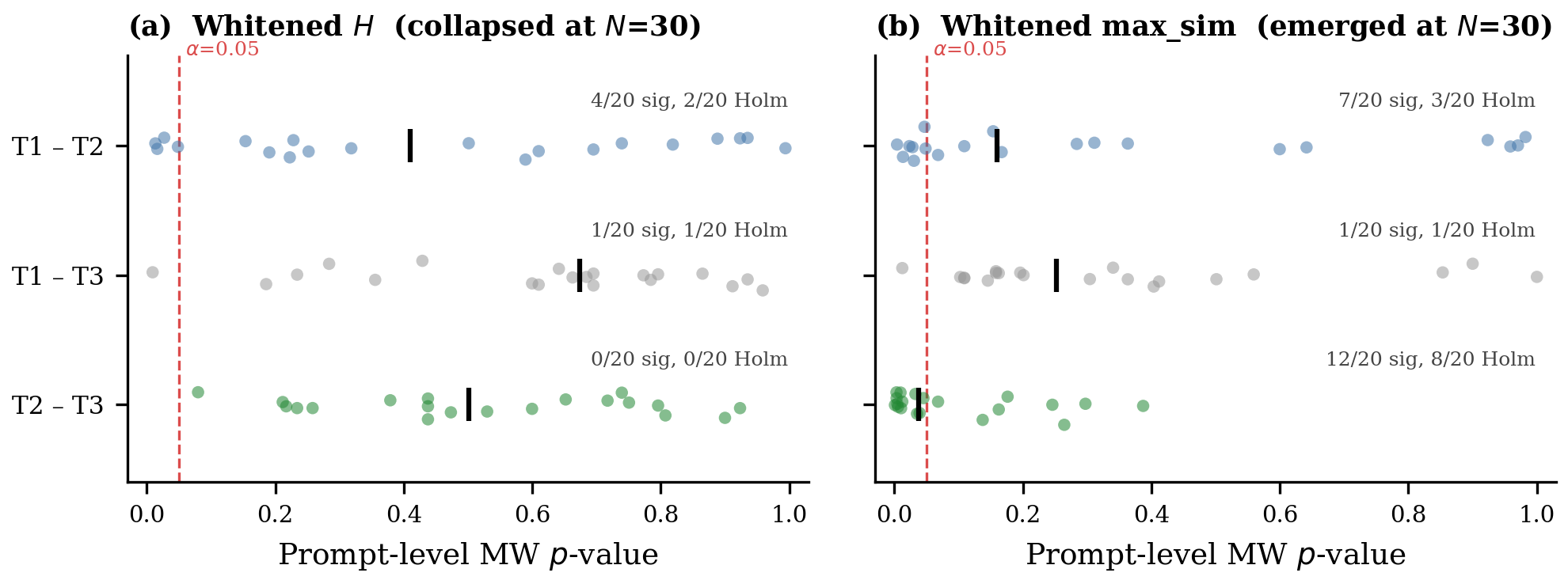}
\caption{The $\Hv$ collapse and max\_sim emergence. Each dot is one seed's prompt-level Mann-Whitney $p$-value (20 seeds total). Vertical ticks: median. Red dashed line: $\alpha = 0.05$. \textbf{(a)}~Whitened $\Hv$ at $N = 30$: $p$-values scattered widely for all pairs. The strongest $N = 15$ signals---T1--T3 (65\% Holm) and T2--T3 (30\% Holm)---collapse to 5\% and 0\% significance respectively. \textbf{(b)}~Whitened max\_sim at $N = 30$: T2--T3 (green) clusters near zero (12/20 sig, 8/20 Holm). T1--T2 (blue) shows a partial cluster below $\alpha$ (7/20 sig, 3/20 Holm)---the emerging signal.}
\label{fig:h_collapse}
\end{figure*}


\section{Cross-Experiment Progression}

Across the three papers, the Type~1/2 question follows a trajectory from consistent null to emerging signal (Table~\ref{tab:progression}).

\begin{table}[t]
\centering
\scriptsize
\setlength{\tabcolsep}{6pt}
\begin{tabular}{llrrr}
\toprule
\textbf{Experiment} & \textbf{Metric} & \textbf{Holm} & \textbf{$r$} & \textbf{Dir} \\
\midrule
Static (Paper~2)          & $\Hv$    & 0/20  & $-0.44^*$ & --- \\
Raw contextual (Paper~2)  & norm     & 0/20  & n.s.      & --- \\
Full-spectrum whitened     & max\_sim & 3/20  & $+0.21$   & 17/20 \\
Best spectral band (tail) & max\_sim & 1/20  & $+0.23$   & 16/20 \\
\bottomrule
\end{tabular}
\caption{Type~1/2 separation across experiments. Paper~2 results from single run; Paper~3 results from 20-seed multi-run. The static result (*$p = 0.042$, uncorrected) does not survive Holm correction. The whitened max\_sim result at $r = +0.21$ with 17/20 directional stability is the first hint of separation---underpowered but sign-consistent.}
\label{tab:progression}
\end{table}

The progression reveals a shift in the informative metric. Papers~1 and~2 used norm and $\Hv$; whitening reveals max\_sim as the metric that directly measures the taxonomy's defining property. The T1--T2 max\_sim hint ($r = +0.21$) is too weak for detection at 124M but is directionally consistent with the taxonomy's prediction that Type~2 should show higher cluster commitment than Type~1. The gap between T2--T3 (max\_sim $\Delta = 0.012$) and T1--T2 (max\_sim $\Delta = 0.008$) quantifies the asymmetry: coverage-gap detection is easier than routing-quality detection because it involves a distributional anomaly, not just a precision distinction.


\section{Implications for Detection}

\subsection{Whitening as Representational Preprocessing}

The results establish whitening as a necessary preprocessing step for geometric hallucination detection in contextual hidden states. Existing detection approaches rely on sampling consistency \citep{manakul2023selfcheckgpt} or output-level confidence \citep{varshney2023stitch}; geometric analysis of internal representations offers a complementary channel. Without whitening, the micro-signal regime ($\Hv \approx 0.985$) renders angular metrics nearly useless. With whitening, max\_sim separates Type~2 from Type~3 at Holm-corrected significance, and the predicted condition ordering becomes visible.

Recent work on representation geometry has shown that targeted interventions in specific directions of hidden-state space can steer model behavior \citep{li2024inference} and that truth-value information organizes along linear subspaces \citep{marks2024geometry}. More broadly, probing hidden representations for linguistic and semantic structure is a well-established methodology \citep{belinkov2022probing, hewitt2019structural}. Our findings extend this picture: hallucination-\textit{type} information is encoded in the cluster structure of whitened representations, and the separating metric is peak alignment (max\_sim), not distributional entropy ($\Hv$). A practical detection pipeline should:

\paragraph{Apply full-spectrum whitening.} This transforms the near-saturated contextual space into one where cluster structure is legible.

\paragraph{Use max\_sim as the primary detector.} Peak cluster alignment directly measures the taxonomy's defining property. The T2--T3 result produces the cleanest statistical profile (inverted pseudoreplication, perfect directional stability).

\paragraph{Retain raw norm as a complementary channel.} Unwhitened norm provides a Type~3 signal independent of whitening (4/20 Holm, direction 19--20/20), offering robustness against whitening failures.

\paragraph{Treat Type~1/2 as unresolved at 124M parameters.} A practical system at this scale should classify tokens as ``coverage gap'' (Type~3) vs.\ ``non-coverage-gap'' (Types~1+2, undifferentiated) unless larger models or fundamentally different features are available.

\subsection{Why the Zone Classifier Continues to Fail}

The confusion matrix mean diagonal across 20 seeds is $0.114 \pm 0.006$, consistent with the monotonic decrease across experiments: 0.288 (static) $\to$ 0.168 (raw contextual) $\to$ 0.114 (whitened). This paradox---the signal gets stronger while classification gets worse---reflects the systematic mismatch between the percentile-based zone classifier (designed for broad-range static embeddings) and the narrow-band geometry of whitened contextual space. The zone thresholds are not recalibrated for this space; the statistical tests with prompt-level aggregation are the primary inference.


\section{Discussion}

\subsection{The Fundamental Asymmetry}

The results support a fundamental asymmetry in what a 124M-parameter model can represent. Work on superposition in small transformers \citep{elhage2022toy} has shown that models at this scale encode rich structure in compressed representations; our findings are consistent, revealing a gradient of cluster commitment even where full discrimination fails. Coverage gaps (Type~3) involve token combinations outside the model's training distribution---a distributional anomaly that produces detectable geometric signatures at any scale. Types~1 and~2 both involve familiar tokens in familiar contexts, differing only in routing quality: whether the model's contextual processing steers toward the correct semantic region (producing coherent text) or an incorrect one (producing locally coherent but contextually wrong text). This is a precision distinction that requires sharper contextual attractors than GPT-2-small provides.

The max\_sim ordering ($\text{T2} > \text{T1} > \text{T3}$) confirms that the model \textit{does} encode a gradient of cluster commitment---the geometric vocabulary exists---but the T1--T2 gap (0.008 in max\_sim) is insufficient for reliable discrimination at $N = 30$. The scaling prediction is specific: at larger models where contextual representations form sharper clusters, this gap should widen. The T1--T2 separation should emerge before the overall detection problem is solved, because it requires only increased precision in the existing geometric vocabulary, not a fundamentally different representation.

\subsection{Prompt-Set Sensitivity}

The $\Hv$ false positive demonstrates a general vulnerability of experiments in the micro-signal regime. When effects live in the fourth decimal place, prompt selection can introduce systematic biases that appear robust across generation seeds (because the same prompts are used in every run) but do not generalize to broader prompt samples. The 20-seed multi-run design catches generation variance but cannot catch prompt-set artifacts---only prompt diversification can. This finding has implications beyond this specific study: any hallucination detection experiment using a fixed, small prompt set should validate against prompt expansion before reporting angular metrics as robust.

\subsection{Spectral Mixing Hypothesis}

The spectral analysis comprehensively rejects the spectral mixing hypothesis. Across 6 bands $\times$ 3 metrics $\times$ 20 seeds, no band produces Type~1/2 separation exceeding 15\% prompt-level significance. The absence is uniform: mid-range bands are dead, the transition band contributes only pseudoreplication artifacts, and even the tail band (which carries the strongest Type~3 effects) fails to separate Types~1 and~2. The non-separation is not a measurement artifact; it is a capacity limitation. This parallels findings in truthfulness probing \citep{burns2023discovering, azaria2023internal}, where targeted extraction succeeds but requires sufficient model capacity.

\subsection{Effect Sizes}

The headline effect size ($r = -0.31$ for T2--T3 max\_sim) is moderate. The 40\% Holm rate means a reviewer could reasonably characterize this as marginal. The defense rests on the full statistical profile: perfect directional stability (20/20), inverted pseudoreplication (genuine prompt-level effect), and the correct theoretical ordering across all three conditions. No single number---not the Holm rate, not the effect size, not the $p$-value---tells the story. The convergent evidence across these independent diagnostics does.


\section{Conclusion}

This paper addresses the Type~1/2 collapse from prior work through PCA-whitening with multi-run stability analysis, and the answer reframes the problem.

First, the theoretically correct separating metric is not entropy but peak cluster alignment. Whitened max\_sim separates Type~2 from Type~3 at 40\% Holm-corrected significance ($r = -0.31$, direction 20/20), with condition means following the taxonomy's predicted ordering $\text{T2} > \text{T1} > \text{T3}$---the first empirical confirmation that the geometric vocabulary of cluster commitment is readable in contextual representations.

Second, the first hint of Type~1/2 separation appears via the same metric (15\% Holm, $r = +0.21$, direction 17/20), underpowered at 124M parameters but directionally stable. This generates a specific capacity prediction: the T1--T2 gap should widen at larger scale.

Third, prompt diversification from 15 to 30 prompts per group eliminates a false positive in whitened entropy, demonstrating that prompt-set sensitivity can produce artifactual results that survive multi-seed validation but not prompt expansion. Spectral decomposition localizes this artifact to the dominant principal components (PCs~1--16), confirming the mechanism.

Together with Papers~1 and~2, this work establishes: the geometric prerequisites exist in static space, the Type~3 signature is detectable in contextual representations through norm, whitening amplifies the vocabulary to reveal max\_sim as the theoretically correct separator with an emerging Type~1/2 hint---and the Type~1/2 frontier requires models beyond 124M parameters.


\section*{Limitations}

Several limitations constrain this work. First, all experiments use GPT-2-small (124M parameters). The capacity hypothesis for Type~1/2 implies that larger models should produce the distinction, but this is untested. Second, the max\_sim T2--T3 effect at 40\% Holm survival is moderate, not robust; a skeptic could characterize this as marginal, though the inverted pseudoreplication ratio and perfect directional stability provide independent support. Third, the T1--T2 hint (15\% Holm, 17/20 direction) is explicitly underpowered---it should be treated as a directional prediction, not a confirmed result. Fourth, clustering uses $k = 40$ (MiniBatchKMeans); the sensitivity to cluster count is not systematically explored, though the signal operates through relative cluster alignment rather than absolute cluster assignments. Fifth, the spectral analysis uses $N = 15$/group due to computational cost (${\sim}24$ hours per 20-seed run); the dominant-band $\Hv$ result at $N = 15$ is known to be the artifact that collapses at $N = 30$ and should be interpreted accordingly. Sixth, the $\Hv$ false positive demonstrates that results in the micro-signal regime are vulnerable to prompt-set artifacts; while we address this through diversification, 30 prompts per type may not fully capture the true prompt distribution. Seventh, prompt-level aggregation with $N = 30$/group limits the minimum achievable $p$-value (${\sim}10^{-4}$), constraining power for small effects.


\section*{Ethics Statement}

This work generates text from GPT-2 under controlled conditions. Some generated outputs contain incoherent, false, or potentially misleading content; this is the expected behavior under study. No human subjects were involved. The research aims to improve understanding of hallucination mechanisms toward safer language model deployment.


\section*{Acknowledgments}

This work was conducted independently without institutional funding or GPU resources. The author thanks Claude (Anthropic) for assistance with computational pipeline development, statistical validation, and manuscript preparation. All scientific hypotheses, experimental design, and interpretive analysis are the author's own.


\bibliography{references}

@article{ji2023survey,
  title={Survey of Hallucination in Natural Language Generation},
  author={Ji, Ziwei and Lee, Nayeon and Frieske, Rita and Yu, Tiezheng and Su, Dan and Xu, Yan and Ishii, Etsuko and Bang, Ye Jin and Madotto, Andrea and Fung, Pascale},
  journal={ACM Computing Surveys},
  volume={55},
  number={12},
  pages={1--38},
  year={2023},
  publisher={ACM}
}

@article{huang2023survey,
  title={A Survey on Hallucination in Large Language Models: Principles, Taxonomy, Challenges, and Open Questions},
  author={Huang, Lei and Yu, Weijiang and Ma, Weitao and Zhong, Weihong and Feng, Zhangyin and Wang, Haotian and Chen, Qianglong and Peng, Weihua and Feng, Xiaocheng and Qin, Bing and Liu, Ting},
  journal={arXiv preprint arXiv:2311.05232},
  year={2023}
}

@article{tonmoy2024comprehensive,
  title={A Comprehensive Survey of Hallucination Mitigation Techniques in Large Language Models},
  author={Tonmoy, S.M Towhidul Islam and Zaman, S M Mehedi and Jain, Vinija and Rani, Anku and Rawte, Vipula and Chadha, Aman and Das, Amitava},
  journal={arXiv preprint arXiv:2401.01313},
  year={2024}
}

@inproceedings{zhang2023siren,
  title={Siren's Song in the {AI} Ocean: A Survey on Hallucination in Large Language Models},
  author={Zhang, Yue and Li, Yafu and Cui, Leyang and Cai, Deng and Liu, Lemao and Fu, Tingchen and Huang, Xinting and Zhao, Enbo and Zhang, Yu and Chen, Yulong and others},
  booktitle={arXiv preprint arXiv:2309.01219},
  year={2023}
}

@article{manakul2023selfcheckgpt,
  title={{SelfCheckGPT}: Zero-Resource Black-Box Hallucination Detection for Generative Large Language Models},
  author={Manakul, Potsawee and Liusie, Adian and Gales, Mark J.F.},
  journal={arXiv preprint arXiv:2303.08896},
  year={2023}
}

@article{varshney2023stitch,
  title={A Stitch in Time Saves Nine: Detecting and Mitigating Hallucinations of {LLMs} by Validating Low-Confidence Generation},
  author={Varshney, Neeraj and Yao, Wenlin and Zhang, Hongming and Chen, Jianshu and Yu, Dong},
  journal={arXiv preprint arXiv:2307.03987},
  year={2023}
}

@article{li2023halueval,
  title={{HaluEval}: A Large-Scale Hallucination Evaluation Benchmark for Large Language Models},
  author={Li, Junyi and Cheng, Xiaoxue and Zhao, Wayne Xin and Nie, Jian-Yun and Wen, Ji-Rong},
  journal={arXiv preprint arXiv:2305.11747},
  year={2023}
}

@article{su2023characterizing,
  title={On the Characterization and Measurement of Hallucinations in Language Models},
  author={Su, Jing and Lu, Chufan and Palacharla, Saranya and Padmanabhan, Deepak},
  journal={arXiv preprint arXiv:2311.01517},
  year={2023}
}

@inproceedings{burns2023discovering,
  title={Discovering Latent Knowledge in Language Models Without Supervision},
  author={Burns, Collin and Ye, Haotian and Klein, Dan and Steinhardt, Jacob},
  booktitle={International Conference on Learning Representations},
  year={2023}
}

@inproceedings{azaria2023internal,
  title={The Internal State of an {LLM} Knows When It's Lying},
  author={Azaria, Amos and Mitchell, Tom},
  booktitle={Findings of the Association for Computational Linguistics: EMNLP 2023},
  year={2023}
}

@article{li2024inference,
  title={Inference-Time Intervention: Eliciting Truthful Answers from a Language Model},
  author={Li, Kenneth and Patel, Oam and Vi{\'e}gas, Fernanda and Pfister, Hanspeter and Wattenberg, Martin},
  journal={Advances in Neural Information Processing Systems},
  volume={36},
  year={2024}
}

@article{marks2024geometry,
  title={The Geometry of Truth: Emergent Linear Structure in Large Language Model Representations of True/False Datasets},
  author={Marks, Samuel and Tegmark, Max},
  journal={arXiv preprint arXiv:2310.06824},
  year={2024}
}

@inproceedings{mu2018allbut,
  title={All-but-the-Top: Simple and Effective Postprocessing for Word Representations},
  author={Mu, Jiaqi and Bhat, Suma and Viswanath, Pramod},
  booktitle={International Conference on Learning Representations},
  year={2018}
}

@inproceedings{ethayarajh2019contextual,
  title={How Contextual are Contextualized Word Representations? {Comparing} the Geometry of {BERT}, {ELMo}, and {GPT-2} Embeddings},
  author={Ethayarajh, Kawin},
  booktitle={Proceedings of the 2019 Conference on Empirical Methods in Natural Language Processing},
  pages={55--65},
  year={2019}
}

@inproceedings{cai2021isotropy,
  title={Isotropy in the Contextual Embedding Space: Clusters and Manifolds},
  author={Cai, Xingyu and Huang, Jiaji and Bian, Yuchen and Church, Kenneth},
  booktitle={International Conference on Learning Representations},
  year={2021}
}

@article{radford2019language,
  title={Language Models are Unsupervised Multitask Learners},
  author={Radford, Alec and Wu, Jeffrey and Child, Rewon and Luan, David and Amodei, Dario and Sutskever, Ilya},
  journal={OpenAI Blog},
  year={2019}
}

@inproceedings{vaswani2017attention,
  title={Attention is All You Need},
  author={Vaswani, Ashish and Shazeer, Noam and Parmar, Niki and Uszkoreit, Jakob and Jones, Llion and Gomez, Aidan N and Kaiser, {\L}ukasz and Polosukhin, Illia},
  booktitle={Advances in Neural Information Processing Systems},
  volume={30},
  year={2017}
}

@article{shannon1948mathematical,
  title={A Mathematical Theory of Communication},
  author={Shannon, Claude E.},
  journal={The Bell System Technical Journal},
  volume={27},
  number={3},
  pages={379--423},
  year={1948}
}

@article{holm1979simple,
  title={A Simple Sequentially Rejective Multiple Test Procedure},
  author={Holm, Sture},
  journal={Scandinavian Journal of Statistics},
  volume={6},
  number={2},
  pages={65--70},
  year={1979}
}

@article{vargha2000critique,
  title={A Critique and Improvement of the {CL} Common Language Effect Size Statistics of {McGraw} and {Wong}},
  author={Vargha, Andr{\'a}s and Delaney, Harold D.},
  journal={Journal of Educational and Behavioral Statistics},
  volume={25},
  number={2},
  pages={101--132},
  year={2000}
}

@inproceedings{sculley2010web,
  title={Web-Scale {K-Means} Clustering},
  author={Sculley, D.},
  booktitle={Proceedings of the 19th International Conference on World Wide Web},
  pages={1177--1178},
  year={2010}
}

@article{belinkov2022probing,
  title={Probing Classifiers: Promises, Shortcomings, and Advances},
  author={Belinkov, Yonatan},
  journal={Computational Linguistics},
  volume={48},
  number={1},
  pages={207--219},
  year={2022}
}

@inproceedings{hewitt2019structural,
  title={A Structural Probe for Finding Syntax in Word Representations},
  author={Hewitt, John and Manning, Christopher D.},
  booktitle={Proceedings of the 2019 Conference of the North {American} Chapter of the Association for Computational Linguistics},
  pages={4129--4138},
  year={2019}
}

@article{elhage2022toy,
  title={Toy Models of Superposition},
  author={Elhage, Nelson and Hume, Tristan and Olsson, Catherine and Schiefer, Nicholas and Henighan, Tom and Kravec, Shauna and Hatfield-Dodds, Zac and Lasenby, Robert and Drain, Dawn and Chen, Carol and others},
  journal={arXiv preprint arXiv:2209.10652},
  year={2022}
}

@article{kadavath2022language,
  title={Language Models (Mostly) Know What They Know},
  author={Kadavath, Saurav and Conerly, Tom and Askell, Amanda and Henighan, Tom and Drain, Dawn and Perez, Ethan and Schiefer, Nicholas and Hatfield-Dodds, Zac and DasSarma, Nova and Tran-Johnson, Eli and others},
  journal={arXiv preprint arXiv:2207.05221},
  year={2022}
}

@article{duan2024llmsknow,
  title={{LLMs} Know More Than They Show: On the Intrinsic Representation of {LLM} Hallucinations},
  author={Duan, Hadas and Antal, Mark and Bhatt, Saurav},
  journal={arXiv preprint arXiv:2410.02707},
  year={2024}
}

@inproceedings{bis2021too,
  title={Too Much in Common: Shifting of Embeddings in Transformer Language Models and Its Implications},
  author={Bis, Daniel and Podkorytov, Maksim and Liu, Xiuwen},
  booktitle={Proceedings of the 2021 Conference of the North {American} Chapter of the Association for Computational Linguistics},
  pages={5117--5130},
  year={2021}
}

@inproceedings{rajaee2021isotropy,
  title={How Does the Pre-trained Language Model Handle Linguistic Knowledge? {A} Case Study of the Anisotropy Problem},
  author={Rajaee, Sara and Pilehvar, Mohammad Taher},
  booktitle={Proceedings of the 16th Conference of the European Chapter of the Association for Computational Linguistics},
  year={2021}
}

@article{author2026geometry,
  title={Detecting {LLM} Hallucinations via Embedding Cluster Geometry: A Three-Type Taxonomy with Measurable Signatures},
  author={Korun, Matic},
  journal={arXiv preprint arXiv:2602.14259},
  year={2026}
}

@article{author2026induction,
  title={From Prerequisites to Predictions: Testing a Geometric Hallucination Taxonomy Through Controlled Induction in {GPT-2}},
  author={Korun, Matic},
  journal={arXiv preprint arXiv:2603.00307},
  year={2026}
}


\appendix

\section{Reproducibility Protocol}
\label{sec:reproducibility}

All experiments were conducted on an Intel Core i7-6700 CPU (3.40\,GHz, 4 cores, 8 threads) with 16\,GB RAM, running Ubuntu Linux. No GPU was used. The pipeline uses PyTorch, Transformers (Hugging Face), scikit-learn, NumPy, SciPy, and matplotlib.

\paragraph{Whitening experiment.} Manual autoregressive generation with KV cache, extracting last-layer last-position hidden states at each step (\texttt{output\_hidden\_states=True}). PCA whitening uses 256 components with regularization $\epsilon = 10^{-5}$. Calibration: 40 prompts $\times$ up to 60 tokens. Clustering: MiniBatchKMeans ($k = 40$, batch 1024, $n_{\text{init}} = 5$, seed 42). Multi-run: 20 generation seeds $\times$ ${\sim}310$\,s each. Total whitened experiment: ${\sim}6{,}450$\,s (${\sim}1.8$\,h).

\paragraph{Spectral experiment.} Full PCA (768 components) on calibration. Six fixed bands with logarithmic spacing. Band-specific whitening uses regularization $\epsilon = 10^{-5}$ and adapted cluster count ($k = \min(40, \text{band\_dim}/2)$, minimum~10). Sliding scan: width 64, step 32, Bonferroni correction across all windows $\times$ 3 metrics. Same generation pipeline as whitening experiment. Multi-run: 20 generation seeds $\times$ ${\sim}4{,}270$\,s each. Total spectral experiment: ${\sim}85{,}490$\,s (${\sim}23.7$\,h).

Code and data are available at: \url{https://github.com/x3mm3x/llm-hallucination-whitening-geometry}.

\section{Spectral Band Results}
\label{sec:spectral_results}

\begin{figure*}[h]
\centering
\includegraphics[width=\textwidth]{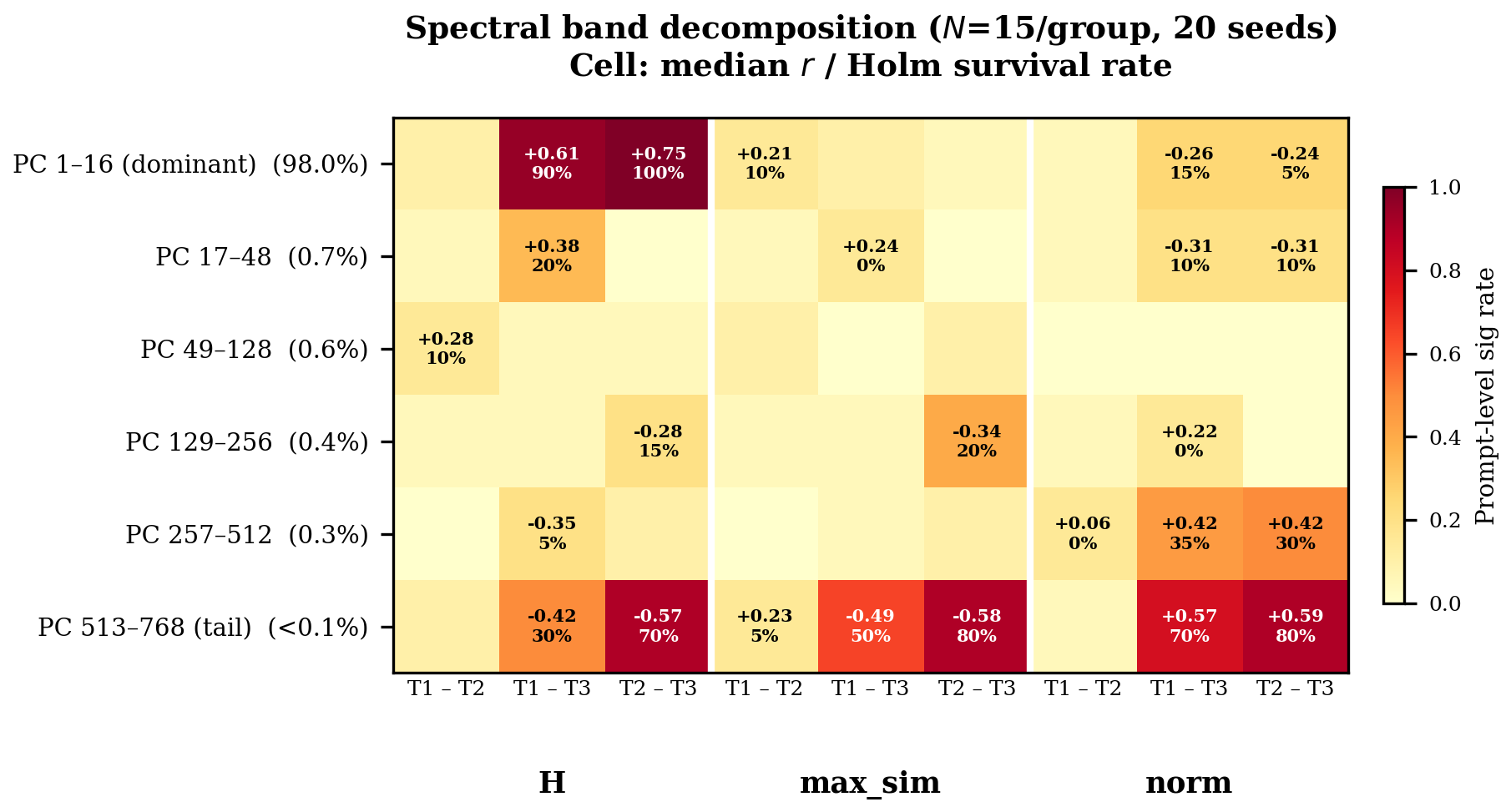}
\caption{Spectral band decomposition ($N = 15$/group, 20 seeds). Cell color: fraction of 20 runs where prompt-level Mann-Whitney $p < 0.05$. Cell annotations (where sig rate $\geq 15$\%): median rank-biserial $r$ (top) and Holm survival rate (bottom). T1--T2 columns are uniformly pale across all bands---Type~1/2 separation does not exist in any spectral region. Dominant band $\Hv$ (PCs~1--16) shows the artifact that collapses at $N = 30$. Tail band effects (PCs~513--768) are genuine but in $<$0.1\% variance dimensions.}
\label{fig:spectral_heatmap}
\end{figure*}

Figure~\ref{fig:spectral_heatmap} presents the full spectral decomposition as a heatmap of prompt-level significance rates across 20 seeds. Each cell shows the median rank-biserial $r$ and Holm survival rate where the significance rate exceeds 15\%. The key patterns are visible: the dominant band (PCs~1--16) carries strong $\Hv$ effects that are the $N = 15$ artifact; the mid-range (PCs~49--256) is essentially uninformative; the tail (PCs~513--768) shows the strongest prompt-robust effects for Type~3 comparisons in near-zero variance dimensions; and the T1--T2 columns are uniformly pale across all bands, confirming the spectral mixing hypothesis is rejected.

\section{Token--Prompt Discordance}
\label{sec:discordance}

\begin{figure}[h]
\centering
\includegraphics[width=\columnwidth]{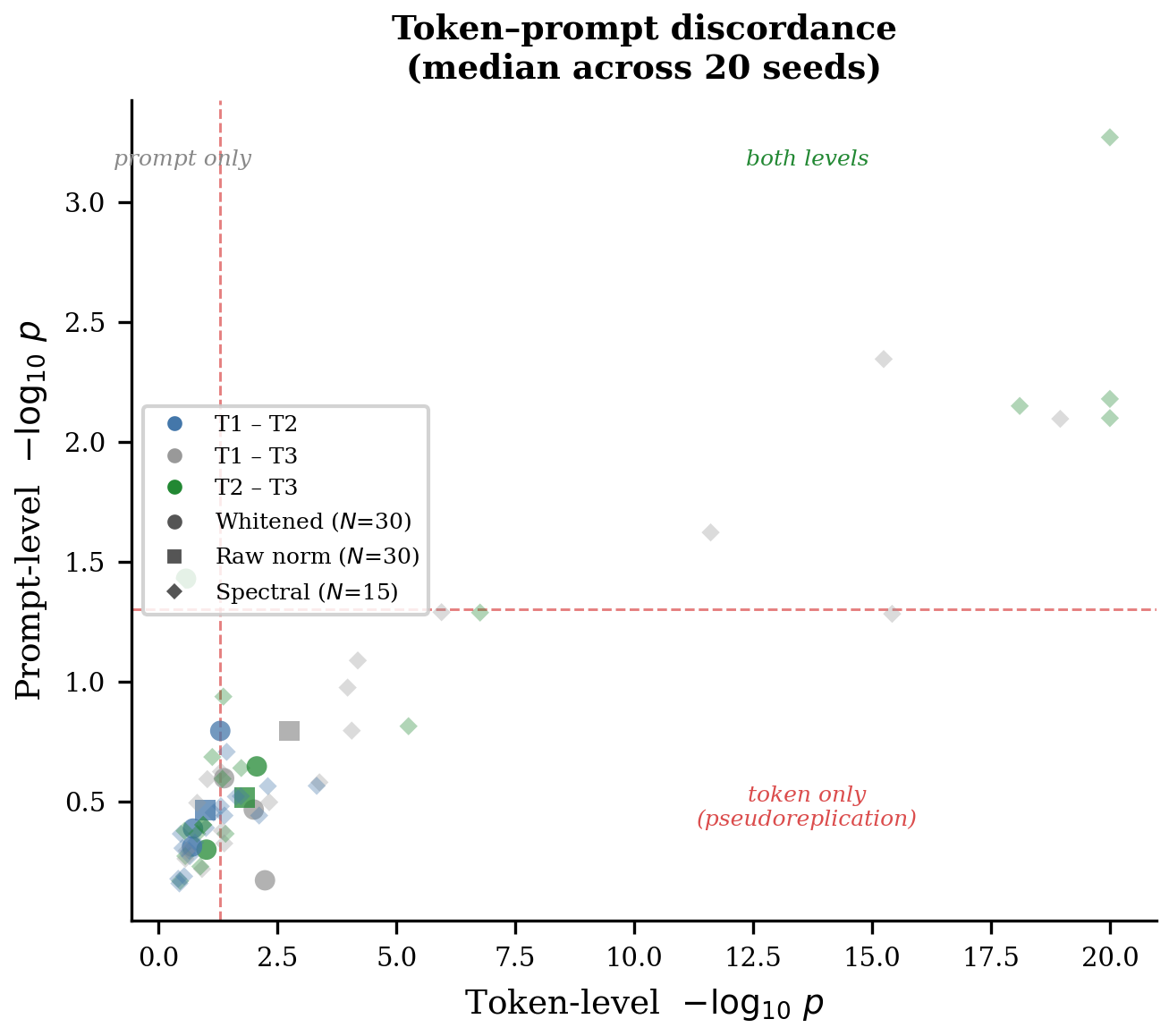}
\caption{Token-level vs.\ prompt-level significance ($-\log_{10} p$, median across 20 seeds) for all experiment--metric--pair combinations. Dashed lines: $p = 0.05$. Circles: whitened ($N = 30$). Squares: raw norm ($N = 30$). Diamonds: spectral bands ($N = 15$). Color: pair type (blue = T1--T2, gray = T1--T3, green = T2--T3). Points in the ``token only'' quadrant (lower right) are pseudoreplication artifacts. The spectral tail produces the most extreme both-levels results.}
\label{fig:discordance}
\end{figure}

Figure~\ref{fig:discordance} plots the median token-level vs.\ prompt-level significance across 20 seeds for every experiment--metric--pair combination. Points in the lower-right quadrant (``token only'') represent pseudoreplication artifacts: high token-level significance that vanishes at prompt level. The upper-right quadrant (``both levels'') contains exclusively Type~3 comparisons from the spectral tail band. Whitened and raw-norm results (circles, squares) cluster near the origin or in the upper-left, while spectral tail results (diamonds) extend to extreme token-level significance.

The max\_sim T2--T3 whitened result (green circle near the $\alpha$ boundary in the upper-left) illustrates the inverted pseudoreplication profile: moderate prompt-level significance with low token-level significance, the opposite of the typical artifact pattern.

\end{document}